\documentclass[akbc,twoside,11pt,lettersize]{article}
\usepackage{misc/akbc}
\usepackage{times}
\usepackage{latexsym}
\usepackage{amsmath}
\usepackage{amssymb}
\usepackage{graphicx}
\usepackage{cleveref}
\usepackage{enumitem}
\usepackage{booktabs}
\usepackage{tabularx}
\usepackage{bbm}

\akbcheading
\ShortHeadings{Artefact Retrieval: Overview of NLP Models with Knowledge Base Access}{Zouhar, Mosbach, Biswas, Klakow}
\finalcopy 

\newcolumntype{L}[1]{>{\raggedright\arraybackslash}m{#1}}
\allowdisplaybreaks

\begin{document}

\title{Artefact Retrieval: Overview of NLP Models\\with Knowledge Base Access}

\author{\name Vilém Zouhar \email vzouhar@lsv.uni-saarland.de \\
       \name Marius Mosbach \email mmosbach@lsv.uni-saarland.de \\
       \name Debanjali Biswas \email dbiswas@coli.uni-saarland.de \\
       \name Dietrich Klakow \email dklakow@lsv.uni-saarland.de \\
       \addr Saarland University, Saarland Informatics Campus, C7 1, R. 001 \\
    66123 Saarbrücken, Germany
   }


\maketitle

\begin{abstract}
Many NLP models gain performance by having access to a knowledge base.
A lot of research has been devoted to devising and improving the way the knowledge base is accessed and incorporated into the model, resulting in a number of mechanisms and pipelines. Despite the diversity of proposed mechanisms, there are patterns in the designs of such systems.
In this paper, we systematically describe the typology of \textit{artefacts} (items retrieved from a knowledge base), retrieval mechanisms and the way these artefacts are \textit{fused} into the model.
This further allows us to uncover combinations of design decisions that had not yet been tried. 
Most of the focus is given to language models, though we also show how question answering, fact-checking and knowledgable dialogue models fit into this system as well.
Having an abstract model which can describe the architecture of specific models also helps with transferring these architectures between multiple NLP tasks.
\end{abstract}

\section{Introduction}

For multiple NLP tasks and primarily language modelling, the BERT \cite{bert}, GPT-2 \cite{radford2019language}, GPT-3 \cite{brown2020language} and T5 \cite{t5} models have seen great success.
Their performance is, however, limited on rare or unseen entities \citep{logan2019baracks, schick2019rare} and knowledge-intensive NLP tasks \citep{chen2020neural}.
As a consequence, they perform poorly on the task of fact-aware language modelling, where the words to be predicted are named entities \citep{logan2019baracks}.

Traditionally question answering systems relied heavily on knowledge base access.
This has been expanded by other NLP tasks for which models have been proposed (\Cref{sec:system-description}) that make use of knowledge bases. 
Not only do they perform better on e.g. fact-aware language modelling, but they can also provide a degree of explainability (which fact was retrieved) and allow for the model and knowledge base components to be trained separately.
The latter is a strong prerequisite for efficient knowledge base manipulation and control, such as removing misinformation or biases \cite{bert-gender, bert-bias-teach} or adding new information. 
This is problematic for models like BERT or T5 because their knowledge is tightly coupled with their generative ability, which is stored implicitly in the parameters.
Such models make use of parametric knowledge representations as opposed to non-parametric ones, which are typically used in retrieval-based approaches (separation of the memory and generation components).
Separating memory from the generative component allows for more flexible architectures, though at the cost of increased complexity.

While using knowledge bases for language modelling is not immediately intuitive, it is very prevalent in the area of open-domain question answering. Commonly in this case the retrieved knowledge (passed to a generative model) is useful for producing an answer to a question in natural language as the model is trained to condition its generation on the retrieved artefact.
Having access to a knowledge base is mandatory for the task of extractive question answering where the output is a span from the available data.
Pre-trained language models have also been shown to be able to perform question-answering in the form of being primed by the question \cite{liu2019multi}. This demonstrates the amount of knowledge that is stored in their parameters, though these models are usually outperformed by models with explicit knowledge base access \cite{kilt}.
Models in the task of slot-filling and fact-checking also benefit from explicit memory access largely because of the same reasons.

Recently there has been an effort in the community to phrase any NLP task in a unified fashion as a sequence to sequence task (seq-2-seq).
Examples of this are phrasing multiple tasks as question answering \citep{mccann2018natural}, the T5 model \citep{t5}, the focus on prompt priming \citep{le2021many, lin2021m6} and GPT3 and its few-shot learning approach for multiple tasks \citep{brown2020language}.
There is also a separate line of research that aims to inject commonsense or factual knowledge into language models by training them on data derived from knowledge graphs.
An example of this is injecting numerical reasoning skills \citep{geva2020injecting} or commonsense knowledge \citep{bosselut2019comet}.
This is orthogonal to retrieval-based approaches (such as dialogue generation with commonsense knowledge base access by \citet{young2018augmenting}) on which we focus in this paper.

\paragraph{Contribution.}

The goal of this paper is to provide a formalism for an abstract model which underlies many specific models for knowledge-intensive tasks with knowledge base access.
Similarly to KILT \cite{kilt}, we hope that this provides a foundation for future research into task-agnostic memory and model architectures.
Lastly, this systematic approach to model description allows us to uncover (1) combinations of mechanisms tried only on one task, which could, however, also work on other NLP tasks and (2) combinations of mechanisms that had not yet been explored for any task.

\paragraph{Outline.}

In \Cref{sec:artefact} we introduce the abstract model underlying other systems together with the different components and the way they can be instantiated. 
The limitations of this schema and also future work with methods not yet explored are discussed in \Cref{sec:discussion}.
We conclude in \Cref{sec:conclusion}.

An important part is also \Cref{sec:system-description} in which different models and approaches for language modelling, question answering, fact-checking and knowledgable dialogue are examined to show how disparate models fit into this schema.

\section{Related Work}

A comprehensive overview of earlier work on neural model-based information retrieval systems together with a general introduction has been done by \citet{mitra2017neural}. 
More recently models such as BERT have been utilized successfully for the task of retrieval itself \citep{nogueira2019multi, soleimani2020bert}.

The aim of KILT \cite{kilt} is to provide a common knowledge base for a number of different NLP tasks (ranging from question answering to fact verification) and to stimulate research in task-agnostic memory architectures. Reformulating various NLP tasks to all use the same knowledge base format provides a stepping stone for the formalisms of artefact retrieval. 

Defining a task-agnostic abstract model is closely related to multi-task learning. The goal of this approach is to improve the performance by training the model on multiple tasks rather than on individual ones \citep{maillard-etal-2021-multi}.
The hope is that representations and generalizations learned for one task will help on another one and vice versa. A strong requirement for this is that the instantiations for different tasks (in the multi-task setup) share significant portions of the model.

An edge-case of this is using a pre-trained BERT model and then fine-tuning it for the new task and/or possibly adding extra layers to match the input and output shapes. Even for BERT, however, it was shown several times \cite{kilt, liu2019multi, sun2019fine, kim2019qe} that training on multiple tasks improves the performance \citep{aghajanyan2021muppet}. This would not be possible without a common model shared among the tasks.
Further related work is discussed in the respective sections when presenting individual NLP models and how they fit into this schema.

\section{Artefact Retrieval} \label{sec:artefact}

\begin{figure*}[t]
    \center
    \includegraphics[width=0.8\textwidth]{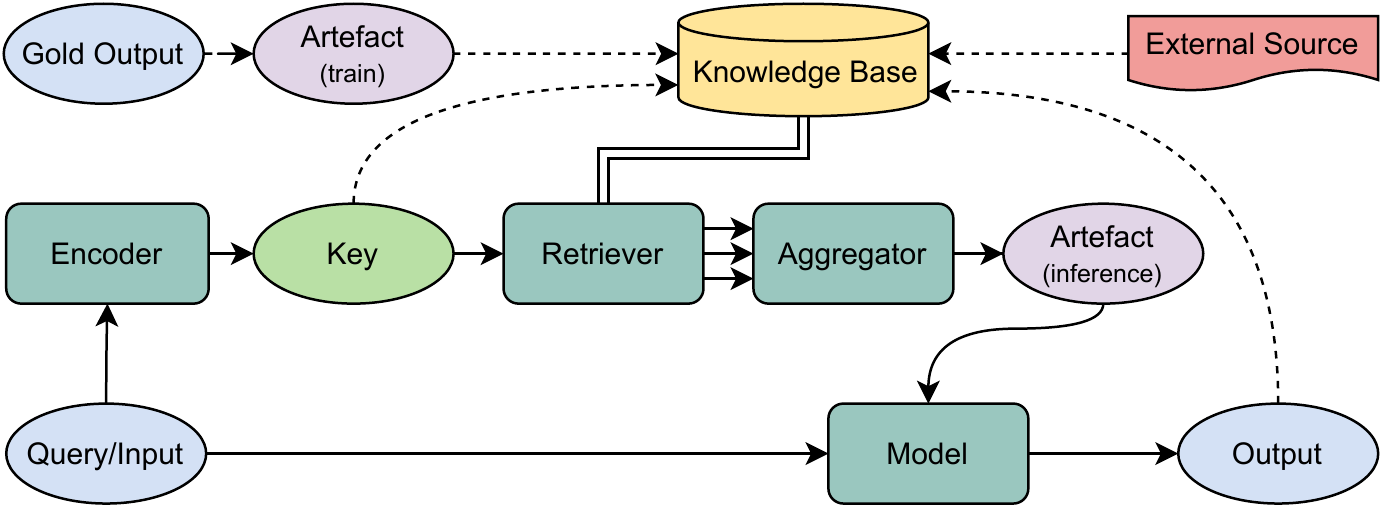}
    \caption{General scheme of NLP models utilizing artefacts by retrieving them from a knowledge base and fusing them into the model in order to produce a better output. Dashed links are utilized only in knowledge base creation and usually not all at once.}
    \label{fig:artefacts_diagram}
\end{figure*}

Systems utilizing knowledge bases usually contain up to four main components (excluding the knowledge base itself): \textbf{encoder}, \textbf{retriever}, \textbf{aggregator} and \textbf{model}. In other works, some of these parts are joined together, most commonly the encoder, the retriever and the aggregator. We separate them for clarity.

A \textit{query} or an \textit{input} is specific to the given task. For language modelling, it is the previous context, for question answering the question, for slot-filling usually the entity and the relation, and for fact-checking the fact to be verified.
In the context of this work, a \textit{knowledge base} is a collection of items, usually (but not necessarily) with a pre-built index that maps keys to values. Prototypically, it is a collection of \textit{documents}, though it can also be a collection of gold training data input-output pairs or a knowledge graph.
\textit{Candidates} are values retrieved from the knowledge base, which may be later post-processed (e.g. reranking or averaging) by the aggregator to form an artefact.
\textit{key} is an object through which the retriever finds suitable artefacts. Commonly the key is a dense vector representation of the input, though it is not necessarily a vector and may be dependent also on an intermediate model computation.
An \textit{artefact} is an object which is (1) dependent on elements retrieved from the knowledge base (e.g. the concatenation of $k$ retrieved documents) and (2) can be used to improve the performance during training and inference. In the simplest example, it is the retrieved value itself, though it can also be multiple retrieved values or their combination.

On a simplified level, models with artefact retrieval usually work the following way, given a query/input $q$ and knowledge base $\mathcal{B}$:
\begin{align*}
& \text{[Key]}~k &=&\quad \texttt{Encoder}(q) \\
& \text{[Candidates]}~C &=&\quad \texttt{Retriever}(\mathcal{B}, k) \\
& \text{[Artefact]}~\xi &=&\quad \texttt{Aggregator}(C) \\
& \text{[Output]}~\hat{y} &=&\quad \texttt{Model}(q, \xi)
\end{align*}

A diagram of this pipeline can be seen in \Cref{fig:artefacts_diagram} (ignore dashed connectors). Pipelines without knowledge base access would only make use of the bottom line (Query/Input $\rightarrow$ Model $\rightarrow$ Output). The aggregator is shown to only depend on the retriever output but in some scenarios may also take the key or the query itself as an input.

\subsection{Artefact Typology}

Most systems used in the literature differ in the encoder, retriever, and model design. We bring attention to four properties that characterize the differences between such systems.

\begin{itemize}[itemsep=-0.24em,topsep=3pt]
\item \textbf{Fusion} (early, late, other)
\item \textbf{Specificity} (sample, task, class)
\item \textbf{KB source} (train, external, dynamic)
\item \textbf{Key \& value type} (dense, sparse, other)
\end{itemize}

\subsection{Fusion} \label{subsec:fusion}

Formally the model estimates $p(y|x,\xi)$ where $y$ is the ground-truth output, $x$ is the query/input and $\xi$ is a retrieved artefact. Fusion \citet{sun2018open} concerns with which point is the artefact made available to the model. It can be presented to the model at the same time as the query/input, e.g. by concatenating $x$ and $\xi$ (early fusion), just before the output is created by the model (late fusion), or somewhere in between. Formally, the model computation is a composition of functions $f_1, \ldots, f_n$. In the simplest example of feed-forward networks, these correspond to single layers and activation functions and on a higher level, they correspond to whole encoder/decoder blocks. The distinction as to what counts as early and late is not clear and for presentation purposes, we consider early fusion at the level of $f_1$ and late fusion at the last stage, $f_n$. These functions themselves may, however, still be composed of multiple others. In \textbf{early fusion}, the artefact is the input together with the query to the first function $f_1$, while in \textbf{late fusion} the query is the single input to $f_1$ and artefact is considered only for $f_n$.
\begin{align*}
& \text{No fusion:} & f_n \circ \ldots \circ f_2 \circ f_1 (q)\\
& \text{Early fusion:} & f_n \circ \ldots \circ f_2 \circ f_1 (q, \xi)\\
& \text{Late fusion:} & f_n ( f_{n-1} \circ \ldots \circ f_1 (q), \xi)\\
&\text{Intermediate fusion:} & f_n \circ \ldots \circ f_k ( f_{k-1} \circ \ldots \circ f_1 (q), \xi)
\end{align*}

Intuitively it makes more sense to prefer early fusion, to maximize the model's access to extra information \citep{izacard2020leveraging, karpukhin2020dense}.
However, this can also be a disadvantage, as the signal from the artefact can get lost during the long computation. In the case of an artefact which is the gold output of a similar query from the training data, later fusion makes more sense. This also allows for a degree of explainability. By examining the forward pass of the last function we could determine what the contribution of the artefact was to the produced output.

\begin{figure*}[ht]
    \center
    \includegraphics[width=0.8\textwidth]{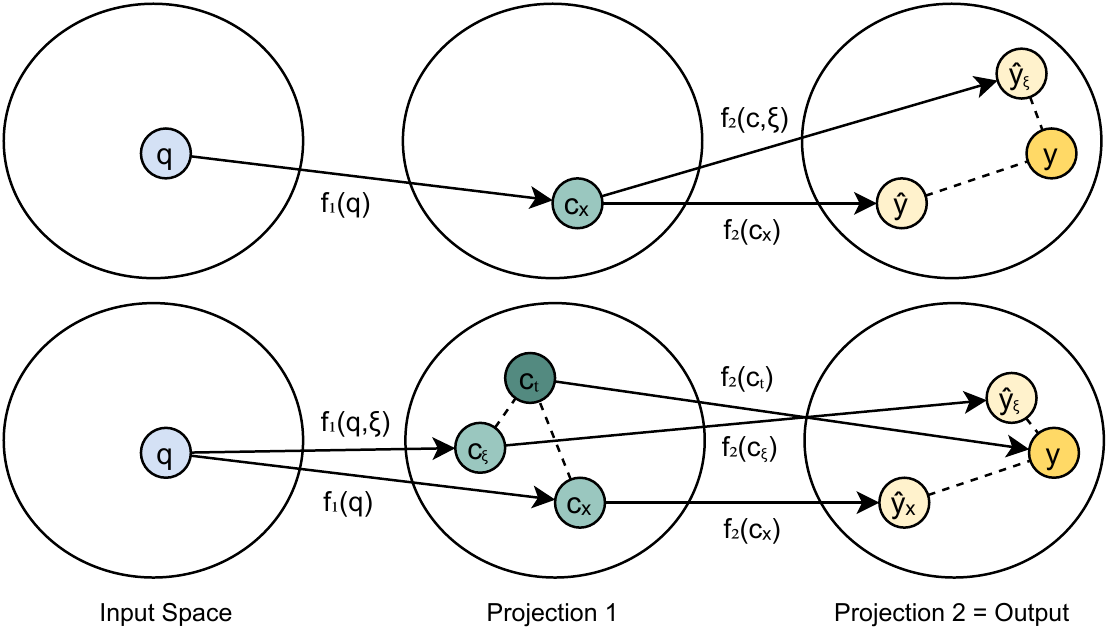}
    \caption{Example of how late (top) and early (bottom) fusions affect the projections into intermediate spaces. The lengths of dashed lines correspond to the loss (longer is greater loss). Adding an artefact $\xi$ to the computation decreases the loss in both early and late fusions.}
    \label{fig:computation_projection}
\end{figure*}

The decision of how late the fusion should be depends heavily on the artefact type.
The application of every function in the chain of computation projects the input to some latent space.
The final function $f_n$ is special because it projects the output of previous functions to the space of possible outputs for the whole model.
In this space, there is the prediction $\hat{y}$ and also the true output $y$.
The task performance metric is defined in this space.
During inference, adding an artefact should ideally move the prediction in the output space closer to the correct output.
Assuming $c$ is the intermediate computation and there are two (overloaded) functions that produce a prediction: $\hat{y}_x = f_n(c)$ and $\hat{y}_\xi = f_n(c,\xi)$.
In circumstances in which adding the artefact helps, $L(\hat{y}_\xi, y) < L(\hat{y}_x, y)$, where $L$ is a loss function such as cross-entropy.
This is illustrated in the first row of \Cref{fig:computation_projection} for $n = 2$.

Assume that we can create an inverse of the last projection and see where the correct output lies in the intermediate representation.
There may be multiple such $c_t: f_n(c_t) = y$ or none, if too much information was lost by the first projection $f_1$.
Further, assume that there is always at least one such $c_t$.
We may then define an intermediate loss $L_i$ for each model computation by measuring the distance of the partial computations to the back-projection.
Similarly to late fusion, we consider two overloaded functions that produce the intermediate representation $c_x = f_1(q)$ and $c_\xi = f_1(q,\xi)$. 
Adding the artefact then ideally moves the intermediate representation closer to the back-projection and reduces the intermediate loss: $L^i(c_\xi, c_t) < L^i(c_x, c_t)$.\footnote{In case of multiple elements that map to $y$, we can define a loss that considers the minimum distance to any of them: ${L^i}' = \min L^i(c, c_t)$. If there is no such element in the projection space, then we may consider the elements that project close to the target $C_t = \arg \min L(f_2(c),y)$.}

This is illustrated in the second row of \Cref{fig:computation_projection}, which depicts a model with only two computational steps: $f_2 \circ f_1$. 
Early fusion (second row) adds the artefact to $f_1$, while late fusion adds it in the next step.
For simplicity in the figure, we consider the standard $L^2$ distance loss between the points.
In both cases, adding the artefact reduced the target loss. For early function, the intermediate loss was also reduced and the target loss was lower.
This does not always happen and complex computations may still at some point project the intermediate computation to the same point regardless of whether an artefact was added earlier or not.
It may also be the case that training with artefacts takes a longer time and the intermediate loss is higher but that the presence of the artefacts will make the model converge to a better optimum (lower generalization error).

Even though the invertibility of projections in this paper is only used as an illustration for the artefact fusion and an intermediate loss does not have to be defined in practice, invertible neural networks are an ongoing topic of research \cite{ardizzone2018analyzing, behrmann2020understanding}.
The combination of artefact retrieval and invertible neural networks has not yet been explored to our knowledge.

\paragraph{Fusion Mechanisms.}

There are common patterns with respect to artefact fusion.
\textit{Priming}, as used by many question-answering systems \cite{guu2020realm,lewis2020retrieval,karpukhin2020dense}, is an example of an early fusion technique. In the simplest scenario, the retrieved paragraphs are prepended to the actual question and put together to the model as one input. 

An extension to this is \textit{key-aware priming}, in which the keys of the retrieved values are also put to the model in an early fusion. The motivation for this in the context of question answering is that the keys may be encoded questions and the retrieved value is simply the answer. Then at inference time, the system can receive a specific question of which the \textit{negation} is stored in the knowledge base. Knowing the answer to the negation of a question may be beneficial in answering the original question, though knowing the question to which the answer corresponds is vital.
Key-aware priming can be also conceptualized as simply storing the key-value pairs as: (key, key+value) and then retrieving the second item of the tuple by the retriever which would also retrieve the key.

For multi-class predictors\footnote{Single-token predictors can be treated as multi-class predictors with the classes being the vocabulary.} like language models it is possible to consider the training data as a knowledge base and when retrieving, suggest k nearest neighbours. This can be followed up by averaging the probabilities of the output classes based on the neighbour distance (performed by the aggregator). For this late fusion approach, \textit{output gating}, the last function $f_n$ can either be a simple convex combination $\hat{y} = \lambda \cdot c + (1-\lambda) \cdot \xi$ or a slightly more complex gating mechanism, which computes $\lambda$ dynamically based on the current input \cite{khandelwal2019generalization, yogatama2021adaptive}. Both are discussed in detail in \Cref{subsec:language-modelling}.

\text{Filtration/masking} can be seen as very late fusion. For slot-filling or fact-aware language modelling, the retriever may provide relevant documents out of which a term incidence vector\footnote{A vector $v$ of the length of the vocabulary filled with ones and zeroes. Then $v_i = 1$ if the $i$-th term occurred in the retrieved documents, otherwise zero.} is constructed. The output of the model is then masked and normalized to prevent outputs unrelated to the query $(\hat{y} = \frac{c\, \odot\, \xi}{|c\, \odot\, \xi|})$.
An example of this is to constrain beam-search decoding to valid names only as done in the recently proposed GENRE model for entity retrieval tasks \citep{de2020autoregressive}.

\subsection{Specificity}

The property of specificity is not directly bound to the pipeline of models using artefact retrieval.
In a wider sense of an artefact being something that improves inference-time performance, the retriever can produce artefacts that help the model, though which are not necessarily dependent on a knowledge base.
In this specific case, the retriever can simply ignore the conditioning on the knowledge base and can be directly fused together with the encoder. In the usual scenario, the encoder makes decisions based on the current query. It may, however, ignore significant parts of the query or be conditioned on a specific task and not take the query into consideration at all.

For the task of open-domain question-answering, retrieving documents based on their embedding's inner product similarity to the query embedding \cite{karpukhin2020dense} is an example of the sample-specific property. 
Presumably significant portions of the query are considered when producing the key for the retriever and different queries produce different artefacts (sets of documents).

Pre-trained language models, such as BERT or GPT\nobreakdash-3, are either fine-tuned on a new task or are primed on prompts either crafted by humans \cite{haley2020bert, misra2020exploring, petroni2019language} or found automatically \cite{shin2020autoprompt, jiang2020can}. These prompts (also called stimuli, constraints, or demonstrations) are then dependent only on the specific task at hand and nothing else. As an example, \citet{radford2019language} affix the prompt \texttt{TL;DR} after a passage to induce summarization behaviour. Crafting task-specific artefact, specifically prompts, is known as \textit{prompt programming} or \textit{prompt engineering} \cite{reynolds2021prompt} and allows the models to perform a wide variety of NLP tasks.

The prompts may be in a structured form with only some parts of it being considered by the encoder. In the case of fact retrieval \cite{shin2020autoprompt}, the input is in the form of (subject, relation). It is then possible to have different prompts based only on the relation (class).
Different conditioning which results in distinct encoder specificities are summarized in the following equations:
\begin{align*}
& \text{Query/input } q = \{q_1, q_2, \ldots q_n\} \hspace{-3cm}\\
& \text{Sample-specific:}  && \texttt{Encoder}(q, \text{task}) \\
& \text{Task-specific:}    && \texttt{Encoder}(\text{task}) \\
& \text{Class-specific:}   && \texttt{Encoder}(q'\subset q, \text{task})
\end{align*}

\subsection{Knowledge Base Source} \label{subsec:kb_source}

The way the knowledge base is created is tightly coupled with its fusion purpose. There are, however, three common underlying patterns that describe the possible creation schemes. In the general architecture overview of pipelines with artefact retrieval in \Cref{fig:artefacts_diagram}, they are depicted with dashed lines.
The most common example is in the task of open-domain question-answering, where the knowledge base, a collection of documents, is a vital component to produce the answer. A collection of documents or knowledge graph structures are examples of knowledge bases populated externally.

A specific contrast to this is having access to the training data during inference. Here, the knowledge base is created as:
\begin{align*}
\mathcal{B} = \{(\texttt{ Encoder}_\text{query}(q), y) | (q, y)\in D_\text{train} \}
\end{align*}

At inference time, the model can then refer back to examples it has already seen with the retriever providing e.g. a weighted combination of nearest neighbours \cite{khandelwal2019generalization} or any other aggregation of found samples. This memorization based approach is usually used with late fusion because, at the end of the computation, the model has access to its proposed output and also to the gold outputs of similar queries. These can be then combined together using e.g. gating.

Finally, the knowledge base can also be created dynamically during inference, such as by keeping a track of already predicted words in language modelling. This knowledge base creation is depicted by the dashed line from \textit{Output} to \textit{Knowledge Base} in \Cref{fig:artefacts_diagram}. This usage of knowledge base lies, however, slightly out of the traditional meaning and understanding of the term.

\subsection{Key \& Value Type}

Commonly the knowledge base is either (1) a collection of candidates (e.g. documents, paragraphs or gold outputs from the training dataset), in which case they are retrieved by similarity to the query or (2) a more structured source of information, such as a knowledge graph.

\paragraph{Vector Space Model.}

Traditionally, question-answering and other NLP systems would use a vector space model for the values (documents) and queries. Such representations usually describe the algorithm for computing the vector representations for both the document and the query. These include standard statistical methods such as TF-IDF \cite{salton1986introduction} or BM25 \cite{robertson1995okapi}. Their disadvantage is that the resulting vectors have the dimension the same as the number of words in the vocabulary. A contrast to this is dense vector representation. LSA \cite{dumais2004latent} is one of the oldest of such methods, but it also includes learned embedding methods based on word2vec \cite{mikolov2013efficient}, doc2vec \cite{le2014distributed}, docT5query \cite{nogueira2019doc2query}, CLSM \cite{shen2014latent} or BERT \cite{bert}. The latter allows for the use of the gradient signal of the latent variable (documents) to fine-tune the index, as done asynchronously by \citet{guu2020realm}.

Splitting the retrieval mechanism into an \texttt{Encoder} and \texttt{Retriever} is not strictly necessary and only follows a common pattern found in many systems. This is usually done for training and speed purposes because e.g. recomputing embeddings for all documents bears too high of a cost for just single retrieval.
In the vector space model, an index (keys for the documents) would be built usually in the pre-processing phase and the knowledge base would constitute a mapping from this index to the documents. Then a vector similarity, usually the cosine similarity, is used which results in the following pipeline:
\begin{align*}
& \mathcal{B}\hspace{-0.25cm}&=&\, \{(\texttt{Encoder}_\text{doc}(d), d) | d\in C \}\hspace{-2cm} \\
& k\hspace{-0.25cm}&=&\, \texttt{ Encoder}_\text{query}(q) & \\
& \xi\hspace{-0.25cm}&=&\, \texttt{ Retriever}(B, k)\, =\, \arg \max_{(v, d) \in B} \text{sim}(k, v) &
\end{align*}

Usually, instead of retrieving just the $\arg \max$, the top-k scoring documents are returned.
When using cosine similarity with normalized vectors, we may substitute the similarity by the inner product $k\, \odot\, v$, which can be approximated efficiently by Maximum Inner Product Search (Approximate Nearest Neighbour Search) algorithms in sublinear query time \cite{johnson2019billion, guo2020accelerating, yang2021linear}.
Instead of documents, which are replaced by paragraphs or spans of texts, depending on the specific work.
It is also possible to store query representation from the training data, as described in \Cref{subsec:kb_source}. The setup is then similar as for document retrieval, with the exception that instead of documents, the retrieved values are gold outputs from the training dataset. The inner product search remains the same.

\paragraph{Knowledge Graphs.}

A vastly different approach has to be chosen when the knowledge base is more structured, such as a knowledge graph. In this case, the knowledge base is usually a directed labelled graph: a set of triples (parent, relation, entity). The parent and the entity are elements of a fixed set of entities and the relation is an element of a fixed set of possible relations (usually orders of magnitudes smaller than the set of entities).

Question-answering over knowledge graphs is a specific, vastly explored \cite{bao2016constraint, lukovnikov2017neural} subfield of non-extractive open-domain question answering.
Popular knowledge graphs, based on Wikipedia, include DBpedia \cite{auer2007dbpedia}, Wikidata \cite{vrandevcic2014wikidata} and YAGO \cite{rebele2016yago}. A comparison of them has been composed by \citet{ringler2017one}. Specific datasets for testing question answering over knowledge graphs are WebQuestions \cite{berant2013semantic} and SimpleQuestions \cite{bordes2015large}.
Knowledge graphs are, however, also used for fact-aware language modelling \cite{logan2019baracks} or fact-checking
\cite{ciampaglia2015computational, tchechmedjiev2019claimskg}.
Simple slot-filling without any reasoning (either multi-source facts or resolving aliases) would be a trivial task. It is then used in the opposite direction, for automatically creating knowledge graphs \cite{yu2014wisdom}.
These knowledge bases found their use even in more distant tasks, such as Word Sense Disambiguation \cite{bevilacqua2020breaking}.
Knowledge graph retrieval is then reduced to finding variables given constraints (a subgraph with free variables that needs to be matched over the knowledge base).

The query construction (which entity and relation should be selected) is handled by the \texttt{Encoder} and is commonly limited to a single restriction (single triplet).
The \texttt{Retriever} is built on top of a database, which stores the knowledge graph. This is vastly faster than MIPS and computing the index in vector space models, but at the cost of a more complicated encoder and constraints to the type of knowledge stored.

\section{Discussion} \label{sec:discussion}

Descriptions of specific systems and approaches for language modelling, knowledge graph language modelling, question answering, fact-checking and knowledgable dialogue in the paradigm of artefact retrieval can be found in \Cref{sec:system-description}.
Summary characteristics of the different models are shown in  \Cref{tab:artefacts_summary}.

\begin{table*}[ht]
\center
\setlength{\tabcolsep}{0.35em} 
\renewcommand{\arraystretch}{1.3} 
\begin{tabular}{p{3.75cm}p{2.1cm}p{1.75cm}p{2.5cm}p{1.5cm}p{1.85cm}}
    \toprule
    Model & Fusion & KB Source & Keys & Values & Aggregation \\
    \midrule
        k-NN LM \newline \begin{minipage}{4cm}\cite{khandelwal2019generalization}\end{minipage}
        & Very late \newline Static convex combination & Train-time & Prefix embd., \newline $L^2$ & Target word & Softmax \\
        Continuous Cache LM \newline \cite{grave2016improving}
        & Very late \newline Static convex combination & Dynamic & Prefix embd., \newline inner product & Target word & Softmax \\
        Dynamic Gating LM \newline \cite{yogatama2021adaptive}
        & Late \newline Dyn. convex \newline combination & Train-time & Prefix encoding, \newline inner product & Target word & Softmax sum \\
        Knowledge Graph LM \newline \cite{logan2019baracks}
        & Intermediate \newline Constraints & External & Entity+relation \newline Discrete struct. & Matching entity & None \\
    \cmidrule{1-1}
        Dense Passage Retrieval\newline \citep{karpukhin2020dense} & Early\newline Input & External & Passage embd.,\newline inner product & Passages & None \\
        Nearest Neighbour QA\newline \citep{lewis2021question} & No model & Train-time & Passage embd.,\newline inner product & Answers & None \\
        CBR-KBQA\newline \citep{das2021case} & Query\newline creation & Train-time\newline External & Query embd.,\newline inner product & Logical forms & New query \\
        PullNet\newline \citep{sun2019pullnet} & Subgraph\newline creation & Multiple\newline External & Entities & Docs and\newline Facts & Iterative \newline join \\
        Universal Schema QA\newline \citep{das2017question} & Intermediate \newline Retrieval & Multiple \newline External & Query embd. \newline Attention & Facts & Iterative \newline projection \\
    \cmidrule{1-1}
        FAKTA\newline \citep{nadeem2019fakta} & Early \newline Input & External \newline Online & Condensed\newline query & Docs & Re-ranking,\newline Filtering\\
    \cmidrule{1-1}
        Wizards of Wikipedia \newline
        \citep{dinan2018wizard} & Intermediate \newline Addition & External & Context+topic,\newline inverted index & Passages & Attention \newline (topic) \\
    \bottomrule
\end{tabular}
\caption{Categorization of described NLP systems in terms of the artefact retrieval typology. \textit{Fusion} describe both where it occurs and what mechanism it employs, \textit{Keys} describes not only the key type but also the retrieval mechanism (e.g. metric).}
\label{tab:artefacts_summary}
\vspace{-0.7cm}
\end{table*}

\subsection{Limitations}

Even though the abstract model presented in this paper is versatile in the description of various models, it may be lacking in subtle ways to describe more complex systems. This complexity may be included in one of the components in order to make it fit, though this obscures clear understanding and comparison across models. An example of this is the implicit aggregation in the form of re-ranking as the final step of the DPR question answering system \citep{karpukhin2020dense}.
We include it in the core model component, though a better understanding of this model would be achieved by adding an additional aggregator element to the abstract model.
This would, however, conflict with most other systems which would not make use of this and for full description, this part of the pipeline would have to be set to simply identity or NOP. This has happened with other models, as documented in \Cref{tab:artefacts_summary}.


\subsection{Future work}

An immediate observation from \Cref{tab:artefacts_summary} is that models for knowledge-intensive tasks, e.g. question answering, do not tend to utilize late fusion nor training-time knowledge base sources yet.
In the opposite directions, experiments could be made with early fusion for language models.
They could also make greater use of knowledge bases with external sources.

Invertible neural networks should be studied in the context of artefact retrieval to determine the exact properties of different fusion mechanisms (e.g. quantifying the discussion in \Cref{subsec:fusion}).

Finally, the area of using multiple separate artefact retrieval pipelines is currently unexplored. This would mean utilizing either (1) multiple knowledge bases with retrieval systems based on the same principle, e.g. dense vector representations or (2) a vastly different knowledge bases with separate retrieval systems, e.g. one with dense vector representation and the other with a knowledge graph.

\section{Summary} \label{sec:conclusion}

In this paper, we presented an abstract model which can describe various systems for NLP tasks utilizing knowledge bases.
The pipeline consists of multiple components: encoder, retriever, aggregator and the core model, typically a generative model, itself.
The abstract model description leads to several key characteristics: fusion, specificity, knowledge base source and key \& value type which share important properties across approaches to model design.

In the Appendix, we showed how several increasingly complex language modelling models proposed in recent years can be described in this paradigm. This is followed by model descriptions in the context of question answering, fact-checking and knowledgeable dialogue.

\acks{
This work was funded by the Deutsche Forschungsgemeinschaft (DFG, German Research Foundation) – Project-ID 232722074 – SFB 1102.
}

\bibliography{misc/bibliography.bib}

\begin{thebibliography}{91}
\providecommand{\natexlab}[1]{#1}
\providecommand{\url}[1]{\texttt{#1}}
\expandafter\ifx\csname urlstyle\endcsname\relax
  \providecommand{\doi}[1]{doi: #1}\else
  \providecommand{\doi}{doi: \begingroup \urlstyle{rm}\Url}\fi

\bibitem[Aamodt and Plaza(1994)]{aamodt1994case}
Agnar Aamodt and Enric Plaza.
\newblock Case-based reasoning: Foundational issues, methodological variations,
  and system approaches.
\newblock \emph{AI communications}, 7\penalty0 (1):\penalty0 39--59, 1994.

\bibitem[Aghajanyan et~al.(2021)Aghajanyan, Gupta, Shrivastava, Chen,
  Zettlemoyer, and Gupta]{aghajanyan2021muppet}
Armen Aghajanyan, Anchit Gupta, Akshat Shrivastava, Xilun Chen, Luke
  Zettlemoyer, and Sonal Gupta.
\newblock Muppet: Massive multi-task representations with pre-finetuning.
\newblock \emph{arXiv preprint arXiv:2101.11038}, 2021.

\bibitem[Ardizzone et~al.(2018)Ardizzone, Kruse, Wirkert, Rahner, Pellegrini,
  Klessen, Maier-Hein, Rother, and K{\"o}the]{ardizzone2018analyzing}
Lynton Ardizzone, Jakob Kruse, Sebastian Wirkert, Daniel Rahner, Eric~W
  Pellegrini, Ralf~S Klessen, Lena Maier-Hein, Carsten Rother, and Ullrich
  K{\"o}the.
\newblock Analyzing inverse problems with invertible neural networks.
\newblock \emph{arXiv preprint arXiv:1808.04730}, 2018.

\bibitem[Auer et~al.(2007)Auer, Bizer, Kobilarov, Lehmann, Cyganiak, and
  Ives]{auer2007dbpedia}
S{\"o}ren Auer, Christian Bizer, Georgi Kobilarov, Jens Lehmann, Richard
  Cyganiak, and Zachary Ives.
\newblock Dbpedia: A nucleus for a web of open data.
\newblock In \emph{The semantic web}, pages 722--735. Springer, 2007.

\bibitem[Baevski and Auli(2018)]{baevski2018adaptive}
Alexei Baevski and Michael Auli.
\newblock Adaptive input representations for neural language modeling.
\newblock \emph{arXiv preprint arXiv:1809.10853}, 2018.

\bibitem[Bahl et~al.(1983)Bahl, Jelinek, and Mercer]{bahl1983maximum}
Lalit~R Bahl, Frederick Jelinek, and Robert~L Mercer.
\newblock A maximum likelihood approach to continuous speech recognition.
\newblock \emph{IEEE transactions on pattern analysis and machine
  intelligence}, \penalty0 (2):\penalty0 179--190, 1983.

\bibitem[Bao et~al.(2016)Bao, Duan, Yan, Zhou, and Zhao]{bao2016constraint}
Junwei Bao, Nan Duan, Zhao Yan, Ming Zhou, and Tiejun Zhao.
\newblock Constraint-based question answering with knowledge graph.
\newblock In \emph{Proceedings of COLING 2016, the 26th International
  Conference on Computational Linguistics: Technical Papers}, pages 2503--2514,
  2016.

\bibitem[Behrmann et~al.(2021)Behrmann, Vicol, Wang, Grosse, and
  Jacobsen]{behrmann2020understanding}
Jens Behrmann, Paul Vicol, Kuan-Chieh Wang, Roger Grosse, and J{\"o}rn-Henrik
  Jacobsen.
\newblock Understanding and mitigating exploding inverses in invertible neural
  networks.
\newblock In \emph{International Conference on Artificial Intelligence and
  Statistics}, pages 1792--1800. PMLR, 2021.

\bibitem[Berant et~al.(2013)Berant, Chou, Frostig, and
  Liang]{berant2013semantic}
Jonathan Berant, Andrew Chou, Roy Frostig, and Percy Liang.
\newblock Semantic parsing on freebase from question-answer pairs.
\newblock In \emph{Proceedings of the 2013 conference on empirical methods in
  natural language processing}, pages 1533--1544, 2013.

\bibitem[Bevilacqua and Navigli(2020)]{bevilacqua2020breaking}
Michele Bevilacqua and Roberto Navigli.
\newblock Breaking through the 80\% glass ceiling: Raising the state of the art
  in word sense disambiguation by incorporating knowledge graph information.
\newblock In \emph{Proceedings of the 58th Annual Meeting of the Association
  for Computational Linguistics}, pages 2854--2864, 2020.

\bibitem[Bhardwaj et~al.(2020)Bhardwaj, Majumder, and Poria]{bert-gender}
Rishabh Bhardwaj, Navonil Majumder, and Soujanya Poria.
\newblock Investigating gender bias in bert, 2020.

\bibitem[Bordes et~al.(2013)Bordes, Usunier, Garcia-Duran, Weston, and
  Yakhnenko]{bordes2013translating}
Antoine Bordes, Nicolas Usunier, Alberto Garcia-Duran, Jason Weston, and Oksana
  Yakhnenko.
\newblock Translating embeddings for modeling multi-relational data.
\newblock \emph{Advances in neural information processing systems}, 26, 2013.

\bibitem[Bordes et~al.(2015)Bordes, Usunier, Chopra, and
  Weston]{bordes2015large}
Antoine Bordes, Nicolas Usunier, Sumit Chopra, and Jason Weston.
\newblock Large-scale simple question answering with memory networks.
\newblock \emph{arXiv preprint arXiv:1506.02075}, 2015.

\bibitem[Bosselut et~al.(2019)Bosselut, Rashkin, Sap, Malaviya, Celikyilmaz,
  and Choi]{bosselut2019comet}
Antoine Bosselut, Hannah Rashkin, Maarten Sap, Chaitanya Malaviya, Asli
  Celikyilmaz, and Yejin Choi.
\newblock Comet: Commonsense transformers for automatic knowledge graph
  construction.
\newblock In \emph{Proceedings of the 57th Annual Meeting of the Association
  for Computational Linguistics}, pages 4762--4779, 2019.

\bibitem[Brown et~al.(2020)Brown, Mann, Ryder, Subbiah, Kaplan, Dhariwal,
  Neelakantan, Shyam, Sastry, Askell, Agarwal, Herbert-Voss, Krueger, Henighan,
  Child, Ramesh, Ziegler, Wu, Winter, Hesse, Chen, Sigler, Litwin, Gray, Chess,
  Clark, Berner, McCandlish, Radford, Sutskever, and Amodei]{brown2020language}
Tom~B. Brown, Benjamin Mann, Nick Ryder, Melanie Subbiah, Jared Kaplan,
  Prafulla Dhariwal, Arvind Neelakantan, Pranav Shyam, Girish Sastry, Amanda
  Askell, Sandhini Agarwal, Ariel Herbert-Voss, Gretchen Krueger, Tom Henighan,
  Rewon Child, Aditya Ramesh, Daniel~M. Ziegler, Jeffrey Wu, Clemens Winter,
  Christopher Hesse, Mark Chen, Eric Sigler, Mateusz Litwin, Scott Gray,
  Benjamin Chess, Jack Clark, Christopher Berner, Sam McCandlish, Alec Radford,
  Ilya Sutskever, and Dario Amodei.
\newblock Language models are few-shot learners.
\newblock 2020.

\bibitem[Chen(2020)]{chen2020neural}
Charles~L Chen.
\newblock \emph{Neural Network Models for Tasks in Open-Domain and
  Closed-Domain Question Answering}.
\newblock Ohio University, 2020.

\bibitem[Chen et~al.(2017)Chen, Fisch, Weston, and Bordes]{chen2017reading}
Danqi Chen, Adam Fisch, Jason Weston, and Antoine Bordes.
\newblock Reading wikipedia to answer open-domain questions.
\newblock In \emph{Proceedings of the 55th Annual Meeting of the Association
  for Computational Linguistics (Volume 1: Long Papers)}, pages 1870--1879,
  2017.

\bibitem[Ciampaglia et~al.(2015)Ciampaglia, Shiralkar, Rocha, Bollen, Menczer,
  and Flammini]{ciampaglia2015computational}
Giovanni~Luca Ciampaglia, Prashant Shiralkar, Luis~M Rocha, Johan Bollen,
  Filippo Menczer, and Alessandro Flammini.
\newblock Computational fact checking from knowledge networks.
\newblock \emph{PloS one}, 10\penalty0 (6):\penalty0 e0128193, 2015.

\bibitem[Das et~al.(2017)Das, Zaheer, Reddy, and McCallum]{das2017question}
Rajarshi Das, Manzil Zaheer, Siva Reddy, and Andrew McCallum.
\newblock Question answering on knowledge bases and text using universal schema
  and memory networks.
\newblock \emph{arXiv preprint arXiv:1704.08384}, 2017.

\bibitem[Das et~al.(2021)Das, Zaheer, Thai, Godbole, Perez, Lee, Tan,
  Polymenakos, and McCallum]{das2021case}
Rajarshi Das, Manzil Zaheer, Dung Thai, Ameya Godbole, Ethan Perez, Jay-Yoon
  Lee, Lizhen Tan, Lazaros Polymenakos, and Andrew McCallum.
\newblock Case-based reasoning for natural language queries over knowledge
  bases.
\newblock \emph{arXiv preprint arXiv:2104.08762}, 2021.

\bibitem[De~Cao et~al.(2020)De~Cao, Izacard, Riedel, and
  Petroni]{de2020autoregressive}
Nicola De~Cao, Gautier Izacard, Sebastian Riedel, and Fabio Petroni.
\newblock Autoregressive entity retrieval.
\newblock \emph{arXiv preprint arXiv:2010.00904}, 2020.

\bibitem[Devlin et~al.(2019)Devlin, Chang, Lee, and Toutanova]{bert}
Jacob Devlin, Ming-Wei Chang, Kenton Lee, and Kristina Toutanova.
\newblock Bert: Pre-training of deep bidirectional transformers for language
  understanding, 2019.

\bibitem[Dinan et~al.(2018)Dinan, Roller, Shuster, Fan, Auli, and
  Weston]{dinan2018wizard}
Emily Dinan, Stephen Roller, Kurt Shuster, Angela Fan, Michael Auli, and Jason
  Weston.
\newblock Wizard of wikipedia: Knowledge-powered conversational agents.
\newblock \emph{arXiv preprint arXiv:1811.01241}, 2018.

\bibitem[Dodge et~al.(2015)Dodge, Gane, Zhang, Bordes, Chopra, Miller, Szlam,
  and Weston]{dodge2015evaluating}
Jesse Dodge, Andreea Gane, Xiang Zhang, Antoine Bordes, Sumit Chopra, Alexander
  Miller, Arthur Szlam, and Jason Weston.
\newblock Evaluating prerequisite qualities for learning end-to-end dialog
  systems.
\newblock \emph{arXiv preprint arXiv:1511.06931}, 2015.

\bibitem[Dumais(2004)]{dumais2004latent}
Susan~T Dumais.
\newblock Latent semantic analysis.
\newblock \emph{Annual review of information science and technology},
  38\penalty0 (1):\penalty0 188--230, 2004.

\bibitem[Geva et~al.(2020)Geva, Gupta, and Berant]{geva2020injecting}
Mor Geva, Ankit Gupta, and Jonathan Berant.
\newblock Injecting numerical reasoning skills into language models.
\newblock In \emph{Proceedings of the 58th Annual Meeting of the Association
  for Computational Linguistics}, pages 946--958, 2020.

\bibitem[Grave et~al.(2016)Grave, Joulin, and Usunier]{grave2016improving}
Edouard Grave, Armand Joulin, and Nicolas Usunier.
\newblock Improving neural language models with a continuous cache.
\newblock \emph{arXiv preprint arXiv:1612.04426}, 2016.

\bibitem[Grave et~al.(2017)Grave, Ciss{\'e}, and Joulin]{grave2017unbounded}
Edouard Grave, Moustapha Ciss{\'e}, and Armand Joulin.
\newblock Unbounded cache model for online language modeling with open
  vocabulary.
\newblock \emph{arXiv preprint arXiv:1711.02604}, 2017.

\bibitem[Guo et~al.(2020)Guo, Sun, Lindgren, Geng, Simcha, Chern, and
  Kumar]{guo2020accelerating}
Ruiqi Guo, Philip Sun, Erik Lindgren, Quan Geng, David Simcha, Felix Chern, and
  Sanjiv Kumar.
\newblock Accelerating large-scale inference with anisotropic vector
  quantization.
\newblock In \emph{International Conference on Machine Learning}, pages
  3887--3896. PMLR, 2020.

\bibitem[Guu et~al.(2020)Guu, Lee, Tung, Pasupat, and Chang]{guu2020realm}
Kelvin Guu, Kenton Lee, Zora Tung, Panupong Pasupat, and Ming-Wei Chang.
\newblock Realm: Retrieval-augmented language model pre-training.
\newblock \emph{arXiv preprint arXiv:2002.08909}, 2020.

\bibitem[Haley(2020)]{haley2020bert}
Coleman Haley.
\newblock This is a bert. now there are several of them. can they generalize to
  novel words?
\newblock In \emph{Proceedings of the Third BlackboxNLP Workshop on Analyzing
  and Interpreting Neural Networks for NLP}, pages 333--341, 2020.

\bibitem[Hochreiter and Schmidhuber(1997)]{hochreiter1997long}
Sepp Hochreiter and J{\"u}rgen Schmidhuber.
\newblock Long short-term memory.
\newblock \emph{Neural computation}, 9\penalty0 (8):\penalty0 1735--1780, 1997.

\bibitem[Izacard and Grave(2020)]{izacard2020leveraging}
Gautier Izacard and Edouard Grave.
\newblock Leveraging passage retrieval with generative models for open domain
  question answering.
\newblock \emph{arXiv preprint arXiv:2007.01282}, 2020.

\bibitem[Ji et~al.(2014)Ji, Nothman, Hachey, et~al.]{ji2014overview}
Heng Ji, Joel Nothman, Ben Hachey, et~al.
\newblock Overview of tac-kbp2014 entity discovery and linking tasks.
\newblock In \emph{Proc. Text Analysis Conference (TAC2014)}, pages 1333--1339,
  2014.

\bibitem[Jiang et~al.(2020)Jiang, Xu, Araki, and Neubig]{jiang2020can}
Zhengbao Jiang, Frank~F Xu, Jun Araki, and Graham Neubig.
\newblock How can we know what language models know?
\newblock \emph{Transactions of the Association for Computational Linguistics},
  8:\penalty0 423--438, 2020.

\bibitem[Johnson et~al.(2019)Johnson, Douze, and J{\'e}gou]{johnson2019billion}
Jeff Johnson, Matthijs Douze, and Herv{\'e} J{\'e}gou.
\newblock Billion-scale similarity search with gpus.
\newblock \emph{IEEE Transactions on Big Data}, 2019.

\bibitem[Karpukhin et~al.(2020)Karpukhin, Oguz, Min, Lewis, Wu, Edunov, Chen,
  and Yih]{karpukhin2020dense}
Vladimir Karpukhin, Barlas Oguz, Sewon Min, Patrick Lewis, Ledell Wu, Sergey
  Edunov, Danqi Chen, and Wen-tau Yih.
\newblock Dense passage retrieval for open-domain question answering.
\newblock In \emph{Proceedings of the 2020 Conference on Empirical Methods in
  Natural Language Processing (EMNLP)}, pages 6769--6781, 2020.

\bibitem[Khandelwal et~al.(2019)Khandelwal, Levy, Jurafsky, Zettlemoyer, and
  Lewis]{khandelwal2019generalization}
Urvashi Khandelwal, Omer Levy, Dan Jurafsky, Luke Zettlemoyer, and Mike Lewis.
\newblock Generalization through memorization: Nearest neighbor language
  models.
\newblock \emph{arXiv preprint arXiv:1911.00172}, 2019.

\bibitem[Kim et~al.(2019)Kim, Lim, Kim, and Na]{kim2019qe}
Hyun Kim, Joon-Ho Lim, Hyun-Ki Kim, and Seung-Hoon Na.
\newblock Qe bert: bilingual bert using multi-task learning for neural quality
  estimation.
\newblock In \emph{Proceedings of the Fourth Conference on Machine Translation
  (Volume 3: Shared Task Papers, Day 2)}, pages 85--89, 2019.

\bibitem[Krishna et~al.(2021)Krishna, Roy, and Iyyer]{krishna2021hurdles}
Kalpesh Krishna, Aurko Roy, and Mohit Iyyer.
\newblock Hurdles to progress in long-form question answering.
\newblock In \emph{Proceedings of the 2021 Conference of the North American
  Chapter of the Association for Computational Linguistics: Human Language
  Technologies}, pages 4940--4957, 2021.

\bibitem[Le and Mikolov(2014)]{le2014distributed}
Quoc Le and Tomas Mikolov.
\newblock Distributed representations of sentences and documents.
\newblock In \emph{International conference on machine learning}, pages
  1188--1196. PMLR, 2014.

\bibitem[Le~Scao and Rush(2021)]{le2021many}
Teven Le~Scao and Alexander~M Rush.
\newblock How many data points is a prompt worth?
\newblock In \emph{Proceedings of the 2021 Conference of the North American
  Chapter of the Association for Computational Linguistics: Human Language
  Technologies}, pages 2627--2636, 2021.

\bibitem[Lee et~al.(2019)Lee, Chang, and Toutanova]{lee2019latent}
Kenton Lee, Ming-Wei Chang, and Kristina Toutanova.
\newblock Latent retrieval for weakly supervised open domain question
  answering.
\newblock In \emph{Proceedings of the 57th Annual Meeting of the Association
  for Computational Linguistics}, pages 6086--6096, 2019.

\bibitem[Lee et~al.(2020)Lee, Li, Wang, Yih, Ma, and Khabsa]{lee2020language}
Nayeon Lee, Belinda~Z Li, Sinong Wang, Wen-tau Yih, Hao Ma, and Madian Khabsa.
\newblock Language models as fact checkers?
\newblock \emph{arXiv preprint arXiv:2006.04102}, 2020.

\bibitem[Lewis et~al.(2020)Lewis, Perez, Piktus, Petroni, Karpukhin, Goyal,
  K{\"u}ttler, Lewis, Yih, Rockt{\"a}schel, et~al.]{lewis2020retrieval}
Patrick Lewis, Ethan Perez, Aleksandara Piktus, Fabio Petroni, Vladimir
  Karpukhin, Naman Goyal, Heinrich K{\"u}ttler, Mike Lewis, Wen-tau Yih, Tim
  Rockt{\"a}schel, et~al.
\newblock Retrieval-augmented generation for knowledge-intensive nlp tasks.
\newblock \emph{arXiv preprint arXiv:2005.11401}, 2020.

\bibitem[Lewis et~al.(2021)Lewis, Stenetorp, and Riedel]{lewis2021question}
Patrick Lewis, Pontus Stenetorp, and Sebastian Riedel.
\newblock Question and answer test-train overlap in open-domain question
  answering datasets.
\newblock In \emph{Proceedings of the 16th Conference of the European Chapter
  of the Association for Computational Linguistics: Main Volume}, pages
  1000--1008, 2021.

\bibitem[Lin et~al.(2021)Lin, Men, Yang, Zhou, Zhang, Wang, Zhou, Tang, and
  Yang]{lin2021m6}
Junyang Lin, Rui Men, An~Yang, Chang Zhou, Yichang Zhang, Peng Wang, Jingren
  Zhou, Jie Tang, and Hongxia Yang.
\newblock M6: Multi-modality-to-multi-modality multitask mega-transformer for
  unified pretraining.
\newblock 2021.

\bibitem[Liu et~al.(2019)Liu, He, Chen, and Gao]{liu2019multi}
Xiaodong Liu, Pengcheng He, Weizhu Chen, and Jianfeng Gao.
\newblock Multi-task deep neural networks for natural language understanding.
\newblock In \emph{Proceedings of the 57th Annual Meeting of the Association
  for Computational Linguistics}, pages 4487--4496, 2019.

\bibitem[Logan et~al.(2019)Logan, Liu, Peters, Gardner, and
  Singh]{logan2019baracks}
Robert Logan, Nelson~F Liu, Matthew~E Peters, Matt Gardner, and Sameer Singh.
\newblock Barack’s wife hillary: Using knowledge graphs for fact-aware
  language modeling.
\newblock In \emph{Proceedings of the 57th Annual Meeting of the Association
  for Computational Linguistics}, pages 5962--5971, 2019.

\bibitem[Lukovnikov et~al.(2017)Lukovnikov, Fischer, Lehmann, and
  Auer]{lukovnikov2017neural}
Denis Lukovnikov, Asja Fischer, Jens Lehmann, and S{\"o}ren Auer.
\newblock Neural network-based question answering over knowledge graphs on word
  and character level.
\newblock In \emph{Proceedings of the 26th international conference on World
  Wide Web}, pages 1211--1220, 2017.

\bibitem[Maillard et~al.(2021)Maillard, Karpukhin, Petroni, Yih, Oguz,
  Stoyanov, and Ghosh]{maillard-etal-2021-multi}
Jean Maillard, Vladimir Karpukhin, Fabio Petroni, Wen-tau Yih, Barlas Oguz,
  Veselin Stoyanov, and Gargi Ghosh.
\newblock Multi-task retrieval for knowledge-intensive tasks.
\newblock In \emph{Proceedings of the 59th Annual Meeting of the Association
  for Computational Linguistics and the 11th International Joint Conference on
  Natural Language Processing (Volume 1: Long Papers)}, pages 1098--1111,
  Online, August 2021. Association for Computational Linguistics.
\newblock \doi{10.18653/v1/2021.acl-long.89}.
\newblock URL \url{https://aclanthology.org/2021.acl-long.89}.

\bibitem[McCann et~al.(2018)McCann, Keskar, Xiong, and
  Socher]{mccann2018natural}
Bryan McCann, Nitish~Shirish Keskar, Caiming Xiong, and Richard Socher.
\newblock The natural language decathlon: Multitask learning as question
  answering.
\newblock \emph{arXiv preprint arXiv:1806.08730}, 2018.

\bibitem[Merity et~al.(2016)Merity, Xiong, Bradbury, and
  Socher]{merity2016pointer}
Stephen Merity, Caiming Xiong, James Bradbury, and Richard Socher.
\newblock Pointer sentinel mixture models.
\newblock \emph{arXiv preprint arXiv:1609.07843}, 2016.

\bibitem[Metz(2019)]{bert-bias-teach}
Cade Metz.
\newblock We teach ai systems everything, including our biases.
\newblock \emph{The New York Times}, 2019.

\bibitem[Mikolov et~al.(2013)Mikolov, Chen, Corrado, and
  Dean]{mikolov2013efficient}
Tomas Mikolov, Kai Chen, Greg Corrado, and Jeffrey Dean.
\newblock Efficient estimation of word representations in vector space.
\newblock \emph{arXiv preprint arXiv:1301.3781}, 2013.

\bibitem[Misra et~al.(2020)Misra, Ettinger, and Rayz]{misra2020exploring}
Kanishka Misra, Allyson Ettinger, and Julia Rayz.
\newblock Exploring bert’s sensitivity to lexical cues using tests from
  semantic priming.
\newblock In \emph{Proceedings of the 2020 Conference on Empirical Methods in
  Natural Language Processing: Findings}, pages 4625--4635, 2020.

\bibitem[Mitra and Craswell(2017)]{mitra2017neural}
Bhaskar Mitra and Nick Craswell.
\newblock Neural models for information retrieval.
\newblock \emph{arXiv preprint arXiv:1705.01509}, 2017.

\bibitem[Nadeem et~al.(2019)Nadeem, Fang, Xu, Mohtarami, and
  Glass]{nadeem2019fakta}
Moin Nadeem, Wei Fang, Brian Xu, Mitra Mohtarami, and James Glass.
\newblock Fakta: An automatic end-to-end fact checking system.
\newblock In \emph{Proceedings of the 2019 Conference of the North American
  Chapter of the Association for Computational Linguistics (Demonstrations)},
  pages 78--83, 2019.

\bibitem[Nie et~al.(2019)Nie, Chen, and Bansal]{nie2019combining}
Yixin Nie, Haonan Chen, and Mohit Bansal.
\newblock Combining fact extraction and verification with neural semantic
  matching networks.
\newblock In \emph{Proceedings of the AAAI Conference on Artificial
  Intelligence}, volume~33, pages 6859--6866, 2019.

\bibitem[Nogueira et~al.(2019{\natexlab{a}})Nogueira, Lin, and
  Epistemic]{nogueira2019doc2query}
Rodrigo Nogueira, Jimmy Lin, and AI~Epistemic.
\newblock From doc2query to doctttttquery.
\newblock \emph{Online preprint}, 2019{\natexlab{a}}.

\bibitem[Nogueira et~al.(2019{\natexlab{b}})Nogueira, Yang, Cho, and
  Lin]{nogueira2019multi}
Rodrigo Nogueira, Wei Yang, Kyunghyun Cho, and Jimmy Lin.
\newblock Multi-stage document ranking with bert.
\newblock \emph{arXiv preprint arXiv:1910.14424}, 2019{\natexlab{b}}.

\bibitem[Petroni et~al.(2019)Petroni, Rockt{\"a}schel, Riedel, Lewis, Bakhtin,
  Wu, and Miller]{petroni2019language}
Fabio Petroni, Tim Rockt{\"a}schel, Sebastian Riedel, Patrick Lewis, Anton
  Bakhtin, Yuxiang Wu, and Alexander Miller.
\newblock Language models as knowledge bases?
\newblock In \emph{Proceedings of the 2019 Conference on Empirical Methods in
  Natural Language Processing and the 9th International Joint Conference on
  Natural Language Processing (EMNLP-IJCNLP)}, pages 2463--2473, 2019.

\bibitem[Petroni et~al.(2020)Petroni, Piktus, Fan, Lewis, Yazdani, De~Cao,
  Thorne, Jernite, Karpukhin, Maillard, et~al.]{kilt}
Fabio Petroni, Aleksandra Piktus, Angela Fan, Patrick Lewis, Majid Yazdani,
  Nicola De~Cao, James Thorne, Yacine Jernite, Vladimir Karpukhin, Jean
  Maillard, et~al.
\newblock Kilt: a benchmark for knowledge intensive language tasks.
\newblock \emph{arXiv preprint arXiv:2009.02252}, 2020.

\bibitem[Radford et~al.(2019)Radford, Wu, Child, Luan, Amodei, and
  Sutskever]{radford2019language}
Alec Radford, Jeffrey Wu, Rewon Child, David Luan, Dario Amodei, and Ilya
  Sutskever.
\newblock Language models are unsupervised multitask learners.
\newblock \emph{OpenAI blog}, 1\penalty0 (8):\penalty0 9, 2019.

\bibitem[Raffel et~al.(2020)Raffel, Shazeer, Roberts, Lee, Narang, Matena,
  Zhou, Li, and Liu]{t5}
Colin Raffel, Noam Shazeer, Adam Roberts, Katherine Lee, Sharan Narang, Michael
  Matena, Yanqi Zhou, Wei Li, and Peter~J. Liu.
\newblock Exploring the limits of transfer learning with a unified text-to-text
  transformer.
\newblock \emph{Journal of Machine Learning Research}, 21\penalty0
  (140):\penalty0 1--67, 2020.
\newblock URL \url{http://jmlr.org/papers/v21/20-074.html}.

\bibitem[Rebele et~al.(2016)Rebele, Suchanek, Hoffart, Biega, Kuzey, and
  Weikum]{rebele2016yago}
Thomas Rebele, Fabian Suchanek, Johannes Hoffart, Joanna Biega, Erdal Kuzey,
  and Gerhard Weikum.
\newblock Yago: A multilingual knowledge base from wikipedia, wordnet, and
  geonames.
\newblock In \emph{International semantic web conference}, pages 177--185.
  Springer, 2016.

\bibitem[Reimers and Gurevych(2020)]{reimers2020curse}
Nils Reimers and Iryna Gurevych.
\newblock The curse of dense low-dimensional information retrieval for large
  index sizes.
\newblock \emph{arXiv preprint arXiv:2012.14210}, 2020.

\bibitem[Reynolds and McDonell(2021)]{reynolds2021prompt}
Laria Reynolds and Kyle McDonell.
\newblock Prompt programming for large language models: Beyond the few-shot
  paradigm.
\newblock In \emph{Extended Abstracts of the 2021 CHI Conference on Human
  Factors in Computing Systems}, pages 1--7, 2021.

\bibitem[Riedel et~al.(2013)Riedel, Yao, McCallum, and
  Marlin]{riedel2013relation}
Sebastian Riedel, Limin Yao, Andrew McCallum, and Benjamin~M Marlin.
\newblock Relation extraction with matrix factorization and universal schemas.
\newblock In \emph{Proceedings of the 2013 Conference of the North American
  Chapter of the Association for Computational Linguistics: Human Language
  Technologies}, pages 74--84, 2013.

\bibitem[Ringler and Paulheim(2017)]{ringler2017one}
Daniel Ringler and Heiko Paulheim.
\newblock One knowledge graph to rule them all? analyzing the differences
  between dbpedia, yago, wikidata \& co.
\newblock In \emph{Joint German/Austrian Conference on Artificial Intelligence
  (K{\"u}nstliche Intelligenz)}, pages 366--372. Springer, 2017.

\bibitem[Robertson et~al.(1995)Robertson, Walker, Jones, Hancock-Beaulieu,
  Gatford, et~al.]{robertson1995okapi}
Stephen~E Robertson, Steve Walker, Susan Jones, Micheline~M Hancock-Beaulieu,
  Mike Gatford, et~al.
\newblock Okapi at trec-3.
\newblock \emph{Nist Special Publication Sp}, 109:\penalty0 109, 1995.

\bibitem[Salton and McGill(1986)]{salton1986introduction}
Gerard Salton and Michael~J McGill.
\newblock Introduction to modern information retrieval.
\newblock 1986.

\bibitem[Schick and Sch{\"u}tze(2020)]{schick2019rare}
Timo Schick and Hinrich Sch{\"u}tze.
\newblock Rare words: A major problem for contextualized embeddings and how to
  fix it by attentive mimicking.
\newblock In \emph{Proceedings of the AAAI Conference on Artificial
  Intelligence}, volume~34, pages 8766--8774, 2020.

\bibitem[Shen et~al.(2014)Shen, He, Gao, Deng, and Mesnil]{shen2014latent}
Yelong Shen, Xiaodong He, Jianfeng Gao, Li~Deng, and Gr{\'e}goire Mesnil.
\newblock A latent semantic model with convolutional-pooling structure for
  information retrieval.
\newblock In \emph{Proceedings of the 23rd ACM international conference on
  conference on information and knowledge management}, pages 101--110, 2014.

\bibitem[Shin et~al.(2020)Shin, Razeghi, Logan~IV, Wallace, and
  Singh]{shin2020autoprompt}
Taylor Shin, Yasaman Razeghi, Robert~L Logan~IV, Eric Wallace, and Sameer
  Singh.
\newblock Autoprompt: Eliciting knowledge from language models with
  automatically generated prompts.
\newblock 2020.

\bibitem[Soleimani et~al.(2020)Soleimani, Monz, and Worring]{soleimani2020bert}
Amir Soleimani, Christof Monz, and Marcel Worring.
\newblock Bert for evidence retrieval and claim verification.
\newblock \emph{Advances in Information Retrieval}, 12036:\penalty0 359, 2020.

\bibitem[Sun et~al.(2019{\natexlab{a}})Sun, Qiu, Xu, and Huang]{sun2019fine}
Chi Sun, Xipeng Qiu, Yige Xu, and Xuanjing Huang.
\newblock How to fine-tune bert for text classification?
\newblock In \emph{China National Conference on Chinese Computational
  Linguistics}, pages 194--206. Springer, 2019{\natexlab{a}}.

\bibitem[Sun et~al.(2018)Sun, Dhingra, Zaheer, Mazaitis, Salakhutdinov, and
  Cohen]{sun2018open}
Haitian Sun, Bhuwan Dhingra, Manzil Zaheer, Kathryn Mazaitis, Ruslan
  Salakhutdinov, and William~W Cohen.
\newblock Open domain question answering using early fusion of knowledge bases
  and text.
\newblock 2018.

\bibitem[Sun et~al.(2019{\natexlab{b}})Sun, Bedrax-Weiss, and
  Cohen]{sun2019pullnet}
Haitian Sun, Tania Bedrax-Weiss, and William Cohen.
\newblock Pullnet: Open domain question answering with iterative retrieval on
  knowledge bases and text.
\newblock In \emph{Proceedings of the 2019 Conference on Empirical Methods in
  Natural Language Processing and the 9th International Joint Conference on
  Natural Language Processing (EMNLP-IJCNLP)}, pages 2380--2390,
  2019{\natexlab{b}}.

\bibitem[Tchechmedjiev et~al.(2019)Tchechmedjiev, Fafalios, Boland, Gasquet,
  Zloch, Zapilko, Dietze, and Todorov]{tchechmedjiev2019claimskg}
Andon Tchechmedjiev, Pavlos Fafalios, Katarina Boland, Malo Gasquet,
  Matth{\"a}us Zloch, Benjamin Zapilko, Stefan Dietze, and Konstantin Todorov.
\newblock Claimskg: A knowledge graph of fact-checked claims.
\newblock In \emph{International Semantic Web Conference}, pages 309--324.
  Springer, 2019.

\bibitem[Thorne et~al.(2018)Thorne, Vlachos, Christodoulopoulos, and
  Mittal]{thorne2018fever}
James Thorne, Andreas Vlachos, Christos Christodoulopoulos, and Arpit Mittal.
\newblock Fever: a large-scale dataset for fact extraction and verification.
\newblock In \emph{Proceedings of the 2018 Conference of the North American
  Chapter of the Association for Computational Linguistics: Human Language
  Technologies, Volume 1 (Long Papers)}, pages 809--819, 2018.

\bibitem[Vaswani et~al.(2017)Vaswani, Shazeer, Parmar, Uszkoreit, Jones, Gomez,
  Kaiser, and Polosukhin]{vaswani2017attention}
Ashish Vaswani, Noam Shazeer, Niki Parmar, Jakob Uszkoreit, Llion Jones,
  Aidan~N Gomez, {\L}ukasz Kaiser, and Illia Polosukhin.
\newblock Attention is all you need.
\newblock In \emph{Advances in neural information processing systems}, pages
  5998--6008, 2017.

\bibitem[Vrande{\v{c}}i{\'c} and Kr{\"o}tzsch(2014)]{vrandevcic2014wikidata}
Denny Vrande{\v{c}}i{\'c} and Markus Kr{\"o}tzsch.
\newblock Wikidata: a free collaborative knowledgebase.
\newblock \emph{Communications of the ACM}, 57\penalty0 (10):\penalty0 78--85,
  2014.

\bibitem[Weston et~al.(2014)Weston, Chopra, and Bordes]{weston2014memory}
Jason Weston, Sumit Chopra, and Antoine Bordes.
\newblock Memory networks.
\newblock \emph{arXiv preprint arXiv:1410.3916}, 2014.

\bibitem[Wu et~al.(2020)Wu, Li, Zhang, Zhou, and Wu]{wu2020topicka}
Sixing Wu, Ying Li, Dawei Zhang, Yang Zhou, and Zhonghai Wu.
\newblock Topicka: Generating commonsense knowledge-aware dialogue responses
  towards the recommended topic fact.
\newblock In \emph{IJCAI}, volume 2020, pages 3766--3772, 2020.

\bibitem[Xu et~al.(2019)Xu, Mohtarami, and Glass]{xu2019adversarial}
Brian Xu, Mitra Mohtarami, and James Glass.
\newblock Adversarial domain adaptation for stance detection.
\newblock \emph{arXiv preprint arXiv:1902.02401}, 2019.

\bibitem[Yang et~al.(2021)Yang, Ren, Shakkottai, Price, Dhillon, and
  Sanghavi]{yang2021linear}
Shuo Yang, Tongzheng Ren, Sanjay Shakkottai, Eric Price, Inderjit~S Dhillon,
  and Sujay Sanghavi.
\newblock Linear bandit algorithms with sublinear time complexity.
\newblock \emph{arXiv preprint arXiv:2103.02729}, 2021.

\bibitem[Yogatama et~al.(2021)Yogatama, de~Masson~d’Autume, and
  Kong]{yogatama2021adaptive}
Dani Yogatama, Cyprien de~Masson~d’Autume, and Lingpeng Kong.
\newblock Adaptive semiparametric language models.
\newblock \emph{Transactions of the Association for Computational Linguistics},
  9:\penalty0 362--373, 2021.

\bibitem[Young et~al.(2018)Young, Cambria, Chaturvedi, Zhou, Biswas, and
  Huang]{young2018augmenting}
Tom Young, Erik Cambria, Iti Chaturvedi, Hao Zhou, Subham Biswas, and Minlie
  Huang.
\newblock Augmenting end-to-end dialogue systems with commonsense knowledge.
\newblock In \emph{Thirty-Second AAAI Conference on Artificial Intelligence},
  2018.

\bibitem[Yu et~al.(2014)Yu, Huang, Cassidy, Ji, Wang, Zhi, Han, Voss, and
  Magdon-Ismail]{yu2014wisdom}
Dian Yu, Hongzhao Huang, Taylor Cassidy, Heng Ji, Chi Wang, Shi Zhi, Jiawei
  Han, Clare Voss, and Malik Magdon-Ismail.
\newblock The wisdom of minority: Unsupervised slot filling validation based on
  multi-dimensional truth-finding.
\newblock In \emph{Proceedings of COLING 2014, the 25th International
  Conference on Computational Linguistics: Technical Papers}, pages 1567--1578,
  2014.

\bibitem[Zaheer et~al.()Zaheer, Guruganesh, Dubey, Ainslie, Alberti, Ontanon,
  Pham, Ravula, Wang, Yang, et~al.]{zaheer2020big}
Manzil Zaheer, Guru Guruganesh, Kumar~Avinava Dubey, Joshua Ainslie, Chris
  Alberti, Santiago Ontanon, Philip Pham, Anirudh Ravula, Qifan Wang, Li~Yang,
  et~al.
\newblock Big bird: Transformers for longer sequences.

\end{thebibliography}
\bibliographystyle{plainnat}

\clearpage

\appendix




\section{System Description} \label{sec:system-description}

This section provides a brief survey of different NLP systems and how their retrieval and fusion mechanisms work in the paradigm of artefact retrieval.
Summaries with respect to this typology are shown in \Cref{tab:artefacts_summary}.

\subsection{Language Modelling} \label{subsec:language-modelling}

The common goal of language modelling is to predict the next word (distribution) given the context of the previous tokens \cite{bahl1983maximum}. The performance is measured by perplexity: the inverse of the geometric average of the whole sequence probability (lower is better). The input to the model is the text prefix and the output of the following word. The prefix is usually the object that is passed to the encoder.

\paragraph{k-Nearest Neighbours.} The language model proposed by \citet{khandelwal2019generalization} utilizes memorization of the training data to decrease the perplexity.
In this approach, the authors first compute representations of all sentence prefixes (keys) and store them together with the following word (values). They then use this for the next word prediction by softmaxing negative $L^2$ distances\footnote{The negative $L^2$ distance is used instead of the common inner product as the similarity measure.
This was shown empirically by the authors to work better.} of $1024$ neighbours.
The representations ($1024$ dimensions) are the output of the last self-attention layer of the trained Transformer-based language model \cite{vaswani2017attention, baevski2018adaptive}.
The importance of the retrieved artefact in the output is determined by the manually set hyperparameter $\lambda \in [0,1]$, resulting in linear (convex) interpolation.

The symbolic working of the model is shown in the following set of equations which is adapted from the original paper (the aggregator is the softmax function; $\text{LM}^{rep.}(X_{<i})$ is the vector representation of the prefix $X_{<i}$ by the trained language model).
\begin{align*}
& \texttt{Encoder:} & k = \textsc{ LM}^{rep.}(X_{<i}) \\
& \texttt{Knowledge base:} & \mathcal{B} = \{ (\texttt{Encoder}(X_{<i}), X_{i}) |X\in D_\text{train}, i < |X| \} \\
& \texttt{Retriever:} & S = \{(r, v) |(r, v) \in \mathcal{N}_{1024}^{L^2}(k)\} \\
& \texttt{Aggregator:} & p_\xi(\hat{X}_i) \propto \hspace{-0.2cm} \sum_{(r,v) \in S} \hspace{-0.2cm} \exp(-||r-k||_2 \cdot v)\\
& \texttt{Model:} & p_m = \lambda \cdot p_\xi + (1-\lambda) \cdot \textsc{LM}(X_{<i})
\end{align*}

The authors also showed that using the training data as a knowledge base outperforms using them for training. This approach was built on the work of \citet{grave2017unbounded} which, however, does not use the state-of-the-art Transformer-based language model but an RNN-based one. Furthermore, they use the inner product (IP) for similarity instead of the $L^2$ distance.

\paragraph{Continuous Cache.}

The previous approach is a continuation of using local vocabulary cache for language modelling, as proposed by \citet{grave2016improving}. The effect of specific history size is examined as well. The usage of cache can be interpreted as using a small local dynamic knowledge base that is being updated after every prediction. This is motivated by the fact that especially rare words tend to occur more probably than by overall uniform distribution, given that they appeared in recent history. The LSTM \cite{hochreiter1997long} hidden state size is again $1024$-dimensional, although this time, it is not used for retrieval but only by the aggregator. For the fusion, the authors propose two methods: (1) linear interpolation, as seen in the previous language model and (2) joint softmax over the artefact and the language model output distribution.

The simplest of the two proposed models (linear combination) is symbolically described in the following equations (the aggregator is the softmax function; $\theta$ is a hyperparameter of the cache distribution).

\begin{align*}
& \texttt{Encoder:} & k = \text{ LSTM}^{hid.}(X_{<i}) \\
& \texttt{Retriever:} & S = \bigcup_{n<N} B_{i-n} \\
& \texttt{Aggregator:} & p_\xi(\hat{X}_i) \propto \sum_{(r,v) \in S} \hspace{-0.2cm} \mathbbm{1}_{v=\hat{X}_i}\exp (\theta \cdot k^T r) \\
& \texttt{Model:} & p_m = \lambda \cdot p_\xi + (1-\lambda) \cdot \textsc{LSTM}(X_{<i}) \\
& \texttt{Knowledge base update:} & \mathcal{B}_{i+1} = \{(k, \arg \max \hat{v})\}
\end{align*}

\paragraph{Dynamic Gating.}

The previously described language models use very late fusion, which is controlled by hyperparameter $\lambda$. \citet{yogatama2021adaptive} propose an approach in which the model itself determines this parameter (now a vector) dynamically based on the current sample. The knowledge base (called long-term memory) is constructed in the same way from the training data as in k-Nearest Neighbours LM. They also introduce short-term memory in the model, which is able to attend to extended local context. Another difference is using two different models for the encoder (vanilla transformer) and the language model itself.
The keys are $512$-dimensional vectors and the retriever uses inner product for nearest neighbour lookup.

The following set of equations, adapted from the original paper, describes the behaviour of the model with respect to the long-term (episodic) memory:
\begin{align*}
& \texttt{Encoder:} & k = \text{ Transformer}^{rep.}(X_{<i}) \\
& \texttt{Knowledge base:} & \mathcal{B} = \{ (\texttt{Encoder}(X_{<i}), X_{i}) |X\hspace{-0.08cm} \in\hspace{-0.08cm} D_\text{train}, i < |X| \} \\
& \texttt{Retriever:} & S = \mathcal{N}_{4}^{IP}(k) \\
& \texttt{Aggregator:} & p_\xi(v') = \sum_{(r, v) \in S} \hspace{-0.2cm} \text{ softmax}(v, k) \cdot v' \\
& \texttt{Model:} & g = \sigma(w_g^T k) \\
& & z = (1-g) \odot p_\xi + g \odot \textsc{LM}(X_{<i}) \\
& & p_m = \text{ softmax}(z, W)
\end{align*}

Since the artefact is processed slightly more by the model compared to the other systems which we describe, e.g. \citet{grave2016improving, khandelwal2019generalization}, we classify it as \textit{late} fusion as opposed to \textit{very late} fusion. This dynamic gating was also applied to the cache knowledge base structure by \citet{merity2016pointer}.

\paragraph{Knowledge Graph LM.}

All of the previous models utilized train-time knowledge base creation with keys representing the contexts and values of the next word predictions. \citet{logan2019baracks} utilize knowledge graphs to specifically increase language modelling performance on named entities. We only describe the simplified retrieval component and omit the details, such as entity rendering and aliases.

At every position (for every query), the model makes a decision as to what type the following token will be: (1) non-entity, (2) unrelated entity or (3) related entity. In the first case, the token is predicted by the standard language model. The other two, however, utilize the knowledge graph access.

Formally there are two knowledge bases used: a static knowledge graph $\mathcal{KG}$ and a local graph $\mathcal{KG}_{<i}$ containing already encountered entities and their relations in the prefix.
The model uses an LSTM unit, the hidden state of which is split into three components $[h_x; h_p; h_r]$ used for: (1) the token type decision, (2) parent entity prediction and (3) relation prediction. When the model makes a decision to predict an unrelated entity, it is sampled by a simple projection to the entity embedding space:
$$\text{softmax}(v_e \cdot (h_p + h_r))$$

The predicted entity $e$ is then added together with its immediate neighbours to the local graph:
$$\mathcal{KG}_{<i+1} = \mathcal{KG}_{<i} \cup \{(e, r, x)|(e, r, x) \in \mathcal{KG}\}$$

In case of a related entity, the model first predicts the parent, then the relation and finally the entity itself. When there are multiple entities in the local graph matching the restriction of the parent and the relation, it is sampled at random. The following set of equations describes the model behaviour in case of a new entity.
\begin{align*}
& \texttt{Encoder:} & p_p(e_p) = \text{ softmax}(v_p \cdot h_p)\\
& & p_r(r)\hspace{0.18cm} = \text{ softmax}(v_r \cdot h_p) \\
& & \hspace{1.1cm} \text{ constrained by } \exists e: (e_p, r, e) \in \mathcal{KG}_{<i}\\
& \texttt{Retriever:} & p_e \underset{\text{\tiny sample}}{\sim} \{(e_p, r, e) | (e_p, r, e) \in \mathcal{KG}_{<i}\} \\
& \texttt{Model:} & p_m = \text{renderer}(p_e) \\
& \texttt{Update knowledge base:} & \mathcal{KG}_{<i+1} = \mathcal{KG}_{<i} \cup \{(e, r, x)|(e, r, x) \in \mathcal{KG}\}
\end{align*}

The main motivation for this approach is factual correctness in language modelling.
Furthermore, it grants a higher degree of explainability and also allows for tighter manipulation and control of the data the model is working with (changing an entity in a relation has a direct impact on the produced output).

\subsection{Question Answering}

There has been a lot of progress in question answering systems. However, the task of long-form question answering was recently shown to be problematic even with state-of-the-art Transformer-based systems \citep{krishna2021hurdles}.
Furthermore, it has been suggested that dense representations are inadequate when scaled up to large index sizes \cite{reimers2020curse}. Despite that, the current trend is based on this retrieval mechanism \citep{lee2019latent,guu2020realm,lewis2020retrieval}.

\paragraph{Dense Passage Retrieval.}

We focus on DPR \citep{karpukhin2020dense} which describes a prototypical question-answering system built with knowledge base access and dense document embeddings.

DPR uses two BERT models to compute the document and question embeddings at the \texttt{[CLS]} token (768-dimensional). They are then fine-tuned so that the inner product (or $L^2$  distance or cosine similarity) of these two vectors is a good measure for document relevancy.
The retrieved passages are reranked by a combination of the original similarity and the BM25 model \citep{robertson1995okapi}.
For every retrieved passage, the probability of a span (up to fixed maximum length) being selected is computed as the product of two tokens (representation from another BERT model) being the start and end ones, respectively. 
Reranking of the answers is done implicitly by choosing the span with the highest probability across all spans.
The maximum similarity search is approximated to make it computationally feasible using FAISS \citep{johnson2019billion}.
\begin{align*}
& \texttt{Encoder:} & k = \textsc{Bert}_\text{query}(q)\\
& \texttt{Knowledge base:} & \mathcal{B} = \{ \textsc{Bert}_\text{doc}(d) |d \in \mathcal{KB} \} \\
& \texttt{Retriever:} & C = \arg \max^n_{d\in B} \text{sim}(q, d) \\
& & C' = \text{Reranker}(C) \\
& & \text{score given by: } \textsc{BM25}(q,d) + \lambda \cdot \text{sim}(q,d) \\
& \texttt{Model:} & P_{\text{start},i}(s) = \text{softmax}(C'_i w_\text{start})_s \\
& & P_{\text{end},i}(t) = \text{softmax}(C'_i w_\text{end})_t \\
& & P_{\text{selection}}(i) = \text{softmax}(C' w_\text{selection})_i \\
\end{align*}

For every passage, $P_\text{selection}(i)$ is the score that this passage contains the answer and for every span $P_{\text{start},i}(s) \cdot P_{\text{end},i}(t)$ is the score of a single span ($s$ to $t$) in a passage.
In this specific model, the aggregator simply passes on all the retrieved passages, though some models simply concatenate all passages and pass it to the model as single string input \citep{izacard2020leveraging}.

\paragraph{Nearest Neighbour QA}
The reliance on training data is taken to its extremes by one of the models in \citet{lewis2021question} in their study of overlap (in terms of paraphrases) between test and train datasets for question answering.
In their approach, they simply use nearest neighbours to retrieve the closest paraphrase to the query (using vector space model) to answer the question.
The encoding is done either by the pretrained DPR retriever \citep{karpukhin2020dense} or by TF-IDF.
As a consequence model can easily answer a question granted that the paraphrase is present in the training set.
\begin{align*}
& \texttt{Encoder:} & k = \textsc{DPR}_\text{query}(q)\\
& \texttt{Knowledge base:} & \mathcal{B} = \{ (\textsc{DPR}_\text{query}(d) |d \in \mathcal{KB} \} \\
& \texttt{Retriever:} & o = \xi = \arg \max_{d\in B}\, \langle q, d\rangle
\end{align*}

Although this nearest neighbour model is a very specific corner-case of the artefact retrieval architecture, it can still be accommodated.

\paragraph{CBR-KBQA}
Models utilizing case-based reasoning methods \citep{aamodt1994case} first retrieve similar cases which are then used in synthesising the current answer.
For question answering, such an architecture has been proposed by \citet{das2021case}.
The retrieved similar queries also contain their logical forms (e.g. SQL or a graph query), based on which a logical form for the current query is constructed using a Transformer-based model BigBird \citep{zaheer2020big}.
This logical form is then executed against a symbolic knowledge base and further refined by another component.
The refinement step solves the issue of sparse knowledge bases and aligns each logical query edge 
This is performed either by pre-trained KB embeddings \citep{bordes2013translating} and similarity search or by using surface form similarity of edge names.
\begin{align*}
& \texttt{Encoder:} & k = \textsc{Roberta}_\textsc{Base}(q)\\
& \texttt{Symbolic knowledge base:} & \mathcal{B_S} = \{ (\textsc{Roberta}_\textsc{Base}(d), f) | (d,f) \in \mathcal{KB}_S \} \\
& \texttt{Retriever:} & C = \arg \max^n_{d\in B}\, \langle q, d \rangle \\
& \texttt{Aggregator:} & \hspace{-2cm} \xi = \textsc{BigBird}(q \texttt{[SEP]} C_1^q \texttt{[SEP]} C_1^f \texttt{[SEP]} \ldots \texttt{[SEP]} C_n^q \texttt{[SEP]} C_n^f) \\
& & \xi' = \textsc{Align}(\xi, \mathcal{KB}) \\
& \texttt{Retriever:} & o = \texttt{Execute}(\xi', \mathcal{KB})
\end{align*}

Advantages of this approach include a higher degree of explainability, higher performance on complex compositional questions and, because the model is non-parametric, extendability of the set of schemas and the knowledge base.
It also demonstrates how some architectures may use more the first retrieval to craft a key for a second retrieval.

\paragraph{PullNet}
The usage of multiple knowledge base sources is further developed by \citet{sun2019pullnet}.
In the proposed model, a subgraph is iteratively constructed.
The subgraph starts with containing only the query and in every iteration step, dubbed \emph{pull}, a node is expanded.
The transition from text to a knowledge graph structure is performed using an entity linker \citep{ji2014overview}.
The output of this process is a set of triples (subject, predicate, object).
Given query $q$, a textual knowledge base $\mathcal{KB}$ and a knowledge-graph $\mathcal{KG}$, the high-level simplified working of PullNet can be summarized as:
\begin{align*}
& \texttt{Retriever:} & C_0 = \textsc{Entities}(q)  \\
& & \hspace{-1.5cm}C_{i+1} = C_i \cup \{ \textsc{PullDocs}(e, q, \mathcal{KB}) \cup \textsc{PullFacts}(e, q, \mathcal{KG}) |v\in \textsc{PullNodes}(C_i) \} \\
& \texttt{Model} & o = \textsc{ClassifyAnswer}(C_T)
\end{align*}

The operation \textsc{PullDocs} retrieves the most relevant documents to query $q$ using IDF, given the constraints of the entity $e$ (documents are linked to entities).
The operation \textsc{PullFacts} retrieves facts from the knowledge-graph given the constraint of the entity $e$ being the subject or the object.
The ordering is based on the inner product between a hidden state \textsc{LSTM} representation of the query $q$ and an embedding of the fact relation.

In the context of the artefact retrieval architecture, this model's working can be encapsulated in a single retrieval component that performs multiple knowledge base accesses followed by the model performing classification on top of the resulting subgraph.

\paragraph{Universal Schema}

A similar approach to joining the reasoning capacity of models over structured knowledge bases and the amount of information on the web has been studied by \citet{das2017question}.
Universal Schema \citep{riedel2013relation}, upon which this model is based, is a way to embed knowledge base data universally in a matrix form.
The task for which this is used is called fill-in-the-blank question answering and each answer is a single entity.
Given query input $q$ and the memory $\mathcal{M}$ the computation is as follows:
\begin{align*}
& \texttt{Encoder:} & c_0 = \textsc{BiLSTM}(q) \\
& \texttt{Retriever/Attention:} & c_t = W_t \big( c_{t-1} + W_p \sum_{(k,v)\in \mathcal{M}} (c_{t-1} \cdot k) \, v \big) \\
& \texttt{Model} & o = \arg \max_{e_i \in \mathcal{E}} c_t \cdot e_i
\end{align*}

Due to the use of memory networks \citep{weston2014memory} and iterative attention, the distinction between model computation and retrieval becomes blurred.
The intermediate context $c_t$, created by attending over the memory can be considered an artefact.


\subsection{Fact Checking}

Despite the research in using the masked language modelling capabilities of pre-trained models for fact checking \cite{lee2020language}, traditional methods rely on external knowledge base access to verify claims. This has the advantage of higher explainability, which is essential in the context of fact-checking.
The pipelines then usually consist of evidence retrieval and verification which maps well to our proposed abstract artefact retrieval model.

The release of FEVER: a large-scale dataset for fact extraction and verification \citep{thorne2018fever} introduced a benchmark which was utilized by multiple works, such as by \citet{nie2019combining} or \citet{nadeem2019fakta}.
We describe the components of the latter.

\paragraph{FAKTA}

The query encoder of FAKTA filters out words that are not verbs, nouns or adjectives.
And appends to it named entities from the claim.
The retriever then fetches relevant documents to this query and if none are found, relaxes the query by incrementally omitting the last tokens.
The model uses several sources of knowledge bases, some of them external such as Google, Bing and Yahoo).
This step is followed up by a re-ranker based on important words (determined by their POS tags) from the document title and the claim.
Relevant documents are further filtered by a CNN based on the work of \citet{xu2019adversarial}.
Another CNN network determines the document stance (agree, disagree or discuss).
Given a claim $q$ the pipeline is as follows:
\begin{align*}
& \texttt{Encoder:} & k =  \text{Filter}_\text{N,V,ADJ}(q) \,^\frown\, \text{Filter}_\text{named entity}(q)\\
& \texttt{Retriever:} & C = \textsc{SearchEngine}(k_{<i}) \\
& & \text{(such that $i$ maximal and results non-empty)} \\
& & C' = \textsc{Reranker}(C) \\
& & C'' = \{d | \textsc{CNN}_\text{relevancy}(d) = \text{relevant})\} \\
& \texttt{Model:} & o = \{ \textsc{CNN}_\text{stance}(d) | d \in C'' \}
\end{align*}

The output of the model is a set of relevant documents with labels regarding the claim.
These inferences can be averaged into a single number which describes how likely it is that the claim is factually true.

\subsection{Knowledgable Open Dialogue}

Finally, we focus on open dialogue with utterance generation dependent on a knowledge base. 
This improves the insufficiencies of parametric model memory in open-ended dialogue centred around facts.
It also increases the explainability of each utterance as it can be traced (with various scores) to the source in the knowledge base.
We will focus on the task and model description by \citet{dinan2018wizard} though there have also been previous works by \citet{dodge2015evaluating} or in the context of a commonsense knowledge base by \citet{young2018augmenting, wu2020topicka}.

\paragraph{Wizards of Wikipedia}
In the pipeline proposed by \citet{dinan2018wizard}, the pool of retrieved documents provided by the method of \citet{chen2017reading} in the form of a term matrix and a simple inverted index lookup.
This pool is fixed so that the results are comparable to their experiments with human annotators.
This part could however also be automated and replaced by a more advanced model which could be finetuned.
The artefact in this pipeline is the weighted average of retrieved documents based on the dot product between the encoded representation and the encoded topic.

Given the conversation history $(q_1, q_2, \ldots, q_n)$ and topic $t$ the first model proposed by \citet{dinan2018wizard} can be described as:
\begin{align*}
& \texttt{Encoder:} & k =  \bigcup_{x \in \{q_{n-1}, q_n, t\}}\textsc{TermVector}(x)\\
& \texttt{Retriever:} & C = \textsc{InvertedIndexLookup}_7(k, \mathcal{KB}) \\
& & \hspace{-1cm} C' = \{\textsc{TransformerEnc}(\text{Title}(d) \,\frown\, \text{Paragraphs}(d)_1) | d \in C\} \\
& \texttt{Aggregator} & \xi = \sum_{d'\in C'} d \times (d \cdot \textsc{TransformerEnc}(t) ) \\
& \texttt{Model:} & o = \textsc{TransformerDec}(\textsc{TransformerEnc}(q)+a)
\end{align*}

The next utterance knowledge source is then dependent on the last two utterances and the topic.
The artefact is then added to the context embedding and similarly to the description in \Cref{fig:computation_projection}, it simply moves the current vector in the embedding space, hopefully, in the right direction.
This can be considered intermediate fusion since the artefact has yet to be processed by a Transformer model.

\end{document}